# A new algorithm for shape matching and pattern recognition using dynamic programming

Noreddine GHERABI
Department of Mathematics and
Computer Science,
University Hassan First, LABO LITEN
Settat, Morocco
gherabi@gmail.com

Abdellatif GHERABI
Department. of Physics,
Laval University,
CANADA
abdellatif.gherabi.1@ulaval.ca

Mohamed BAHAJ
Department of Mathematics and
Computer Science,
University Hassan First, LABO LITEN
Settat, Morocco
mohamedbahaj@gmail.com

*Abstract*— We propose a new method for shape recognition and retrieval based on dynamic programming.
Our approach uses the dynamic programming algorithm to compute the optimal score and to find the optimal alignment between two strings. First, each contour of shape is represented by a set of points. After alignment and matching between two shapes, the contours are transformed into a string of symbols and numbers. Finally we find the best alignment of two complete strings and compute the optimal cost of similarity.
In general, dynamic programming has two phases- the forward phase and the backward phase. In the forward phase, we compute the optimal cost for each subproblem. In the backward phase, we reconstruct the solution that gives the optimal cost.
Our algorithm is tested in a database that contains various shapes such as MPEG-7.

## I. INTRODUCTION

In recent years, researchers have extensively studied visual perception and object recognition, Current techniques for object recognition and classification of the shapes are not yet fully satisfactory solutions provider. To recognize an object several properties can be used, such as shape, color, texture and brightness.
Different search techniques were investigated to retrieve shapes from databases. These research techniques used to extract shape descriptors of each shape that is in the database and use these descriptors as indices in the database. Lowe's SIFT descriptor [1] and the descriptor Mikolajczyk, al. [2] are some examples. With regards to the shape representation technique, the Fourier descriptors have been used as representation of shape for several years. In [3], [4], shapes were represented using a Fourier expansion of the function of their tangent angle and their arc length. The lower-order Fourier coefficients were then used to represent the shape. However, shape representation and description is a difficult task. This is because when a 3-D real world object is projected onto a 2-D image plane, one dimension of object information is lost. As a result, the shape extracted from the image only partially represents the projected object. To make the problem even more complex, shape is often corrupted with noise, arbitrary distortion and occlusion.
Various shape descriptors [5,6] have been proposed recently.

The first descriptor contour was introduced by Jain et al. [7]. It was developed to search for images in a database.
Several studies use the contour as pattern recognition. Mohammad Reza Daliri et al [8] have developed a new method for shape recognition and retrieval, in this method the shape descriptor is based on the angles and distances in order to present the shape in a string of symbols.
In the present paper we explore a different approach to object recognition, namely the possibility of representing a shape using a string of symbols and to recognize and retrieve shapes by operations on strings of symbols. This approach has already been explored in the past, but its performance was not fully satisfied.
Our algorithm analyzes the contour of pairs of shapes. Their contours are recovered and represented by a pair of N points. After alignment, our system transforms each contour into a set of symbols and numbers and the system converts these symbols and numbers into strings sequences. After the system compute the optimal cost of similarity between sequences of strings using the technique of dynamic programming,.

## II. AN OVERVIEW OF OUR APPROACH

Our approach to recognition and retrieval of the shape is based on several steps summarized in Fig 1.

The first step is to analyze the contour of the shape to be studied. The contour is retrieved and represented by a set of points N. The cost of the correspondence between the points $p_i$ and $q_j$ of the two shapes is evaluated by the technique of shape context and dynamic programming; this technique is detailed in Section II-B.
After, the two shapes are aligned using Procrustes analysis, summarized in Section II-C. Then each contour is transformed into a string of symbols and numbers (section II-D), using dynamic programming to compute the similarity between the set of symbols by computing the optimal cost for the retrieval and recognition.

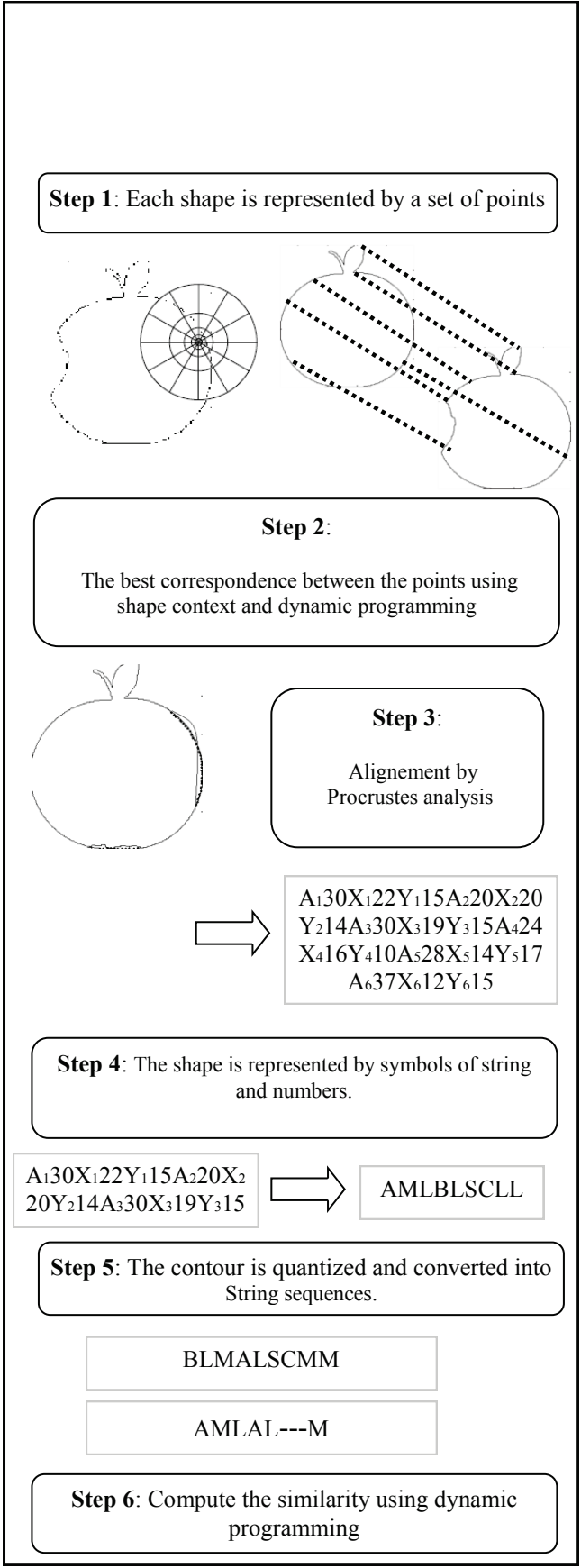

Fig. 1: The algorithm for shape retrieval and recognition.

*A. Matching with Shape Contexts and dynamic programming*

*1. Shape Context*

The Shape Context is a descriptor developed for finding correspondences between point sets which has been introduced by Belongie et al [9, 10, 11]. It is intended to be a way of describing shapes that allows for measuring shape similarity and the recovering of point correspondences. In this approach, a shape is represented by a discrete set of points sampled from the internal or external contours on the shape. These can be obtained as locations of edge pixels as found by an edge detector. The basic idea is to pick N points on the contours of a shape and for every point a log-polar histogram (or the shape context) is computed approximating the distribution of adjacent point locations relative to a reference point. In order to achieve scale invariance, the outer radius for the histograms is set equal to the mean distance between all the pair points

Giving us a set $p = \{p_1, p_2, ...., p_n\} p_i \in \Re^2$ of N points.

Fig. 2.a shows sample points for two shapes.

For a point $p_i$ (*i*=1. . . *N)* on the shape, we compute a coarse histogram **hi** of the relative coordinates of the remaining n-1 points (see Fir.3) :

$$h_i(k) = \#\{q \neq p_i : (q - p_i) \in bin(k)\} \quad (1)$$

This histogram is defined to be the shape context of $p_i$

The bins are uniform in log-polar space, making the descriptor more sensitive to positions of nearby points than those of more distant points. The cost of matching a pair of points $p_i$ and $q_j$ from two shapes is computed as:

$$C(p_i, q_j) = \frac{1}{2} \sum_{k=1}^{K} \frac{[h_i(k) - h_j(k)]^2}{h_i(k) + h_j(k)} \quad (2)$$

Where $h_i(k)$ and $h_j(k)$ denotes the K-bin normalized histogram at $p_i$ and $q_j$ respectively.

*2. Best Matching*

Dynamic programming is the best method for matching a set of points. Thomas B. Sebastian et al [12] presented a novel approach to finding a correspondence between two curves.

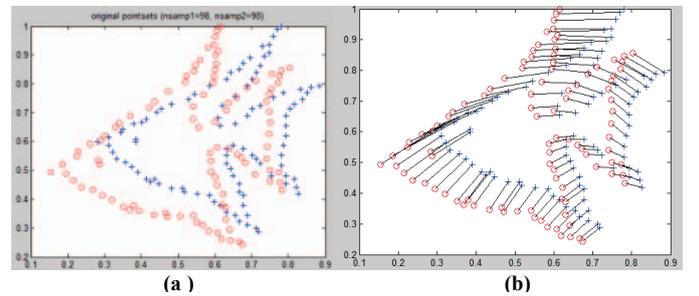

Fig. 2: (a) sampled edge points of two shapes
(b) The Correspondence between points sets using Shape Context costs.

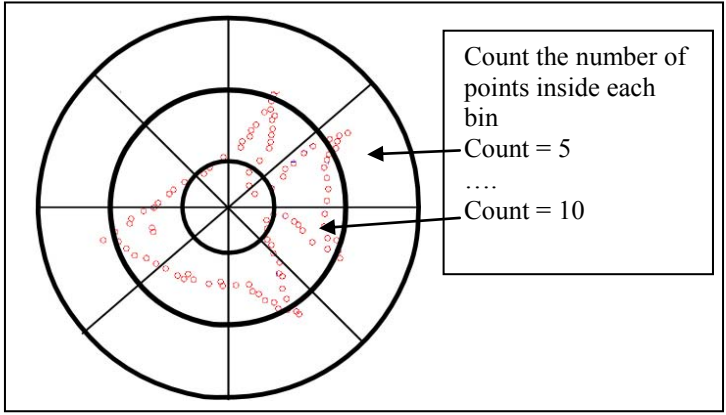

Fig 3: A shape is represented by a discrete set of points sampled regularly along the contours. For every point, a log-polar histogram is computed.

The correspondence is based on a notion of an alignment curve. The optimal correspondence is found by an efficient dynamic programming method.

Given two points $p_i$ in shape A and $q_j$ in shape B, we construct a matrix $C(i,j)$, their values are computed by equation 2. These values are the costs of matching $p_i$ and $q_j$.

Fig. 2.a shows the correspondence between a set of points using Shape Context and dynamic programming.

### B. Alignement by Procrustes analysis

The problem of matching shapes parameterized as a set of points is frequently encountered in the field of imaging. There is usually no problem of determining the correspondences or homologies between the two sets of points. However, when the fixed points were calculated from images, the alignment problem is solved. The Procrustes method is a well known method of alignment of shapes which can still be used to serve at all the homologies between known points in advance. We use Procrustes analysis for aligning shapes. Rangarajan et al[13] present a powerful extension of the Procrustes method to pointsets of differing point counts with unknown correspondences. The result is the softassign Procrustes matching algorithm which iteratively establishes correspondence and rejects non-homologies as outlier.

We use Procrustes analysis only with rotation and translation, because the shapes of different classes are similar.

### C. Shape Descriptor and symbolic representation

In our approach the contour of the shape is represented by a set of points N, each contour is defined by a center of gravity G.

In the first stage, the center of gravity G is located, and then we calculate its maximum distance Dmax with the contour points (resp. The minimum distance Dmin).

Let $x_1$ the contour point such that $Gx_1 = Dmax$, $y_1$ the contour point such that $Gy_1 = Dmin$ and $\mu_1$ is the angle between the lines $(Gx_1)$ and $(Gy_1)$
Then each contour is defined by initial state, this state is characterized by a center of gravity, a long distance (Dmax) and a small distance (Dmin) of the center G and the angle between the lines $(Gx_1)$ and $(Gy_1)$ (See Fig 4).

So the initial state of the contour is:

$$\begin{cases} Distance(Gx_1) = Dmin \\ Distance(Gy_1) = Dmax \\ \mu_1 = Angle(x_1Gy_1) \end{cases} \quad (3)$$

Where, $x_1$ corresponding to $y_1$ and $\mu_1$.

Generally, if we go through all the contour points, for every point $x_i$ in the contour, there are two variables corresponding, the point $y_i$ and angle $\mu_i$.

$$\begin{cases} Distance(Gx_i) = X_i \\ Distance(Gy_i) = Y_i \quad 1 \le i \le N \\ \mu_i = Angle(x_iGy_i) = A_i \end{cases} \quad (4)$$

Then, each state $(x_i, y_i)$ is defined by three variables, (1) the angle $A_i$, (2) distance $X_i$ and (3) distance $Y_i$ (Fig. 5)

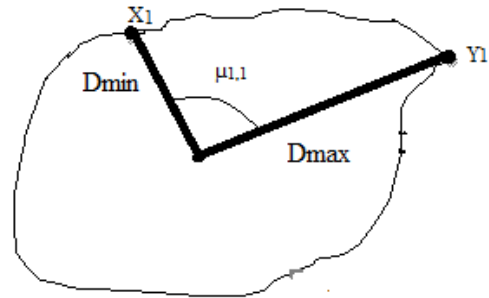

Fig 4. The initial state of the contour.

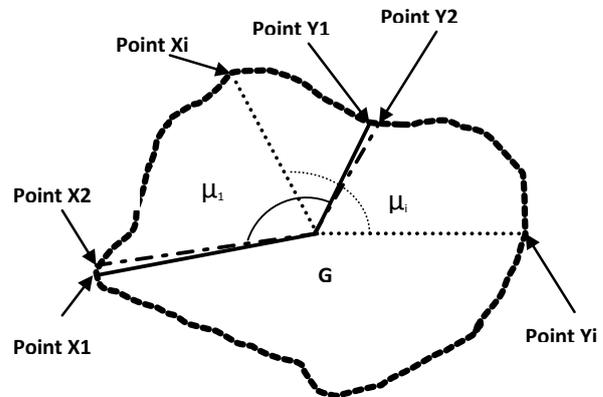

Fig. 5: Transformation of the shape into a symbolic representation. Each state $(x_i, y_i)$ is transformed into three different symbols, one for the angle $\mu_i = x_iGy_i$, the others for two distances from the center of gravity (normalized to $X_i$, and $Y_i$).

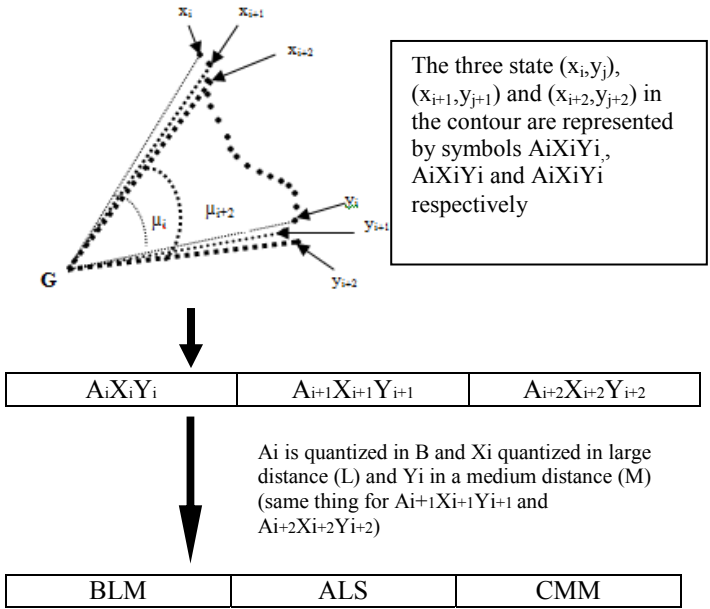

| $A_iX_iY_i$ | $A_{i+1}X_{i+1}Y_{i+1}$ | $A_{i+2}X_{i+2}Y_{i+2}$ |

Ai is quantized in B and Xi quantized in large distance (L) and Yi in a medium distance (M) (same thing for $A_{i+1}X_{i+1}Y_{i+1}$ and $A_{i+2}X_{i+2}Y_{i+2}$)

| BLM | ALS | CMM |

Fig. 6: Converting the contour into String sequences

The distances Xi and Yi are quantified in three bins (S, M, L) according to their values, corresponding to a small, medium and large distance of G. and for each state (xi, yi) we calculate the angle Ai (xiGyi), this angle will be quantized in different K bins between [0, π], ie in K different angles. The value of K is defined in the programming. In our example we have taken K=6, ie six bins (A, B, C, D, E, F). It is also important to observe that by increasing the value of K we obtain a representation at a larger scale, as it will consider angles over more distant points.

Once shape descriptor is created, our algorithm gives a cost for each state according to three variables found ($A_i$, $X_i$ and $Y_i$). These costs will be converted into sequences of strings (see Fig.6 for the scheme of the converting the contour into string sequence).

Since the symbols and numerical values of shape features are reflected in the string sequences, the system can find similar shapes by searching similar string sequences against the query key. Therefore, we apply a dynamic programming approach for string sequence comparisons.

### III. COMPARISON BETWEEN STRINGS SEQUENCES WITH DYNAMIC PROGRAMMING.

After the contours are transformed on the sequence of string, their similarity can be evaluated by an appropriate comparison of the entire string.

The dynamic programming can find the best alignment between two strings with different lengths. When sequences of strings are aligned, sequence alignment scores are computed. The system can find similar sequences by sorting the alignment score.

For this purpose, we propose two methods, namely the Edit Levenstein [14] or algorithm of Needleman-Wunsch[15].

In this paper, we use the algorithm of Needleman/Wunsch . The technique was modified by adding cost of similarity between the symbols.

The algorithm of Needleman-Wunsch is an example of dynamic programming, as the Levenshtein algorithm to which it is related. It guarantees to find the alignment of maximum score. For example, consider two sequences to be aligned, Sequence #1 (represented in the figure 6) and Sequence #2 corresponding to shape A and shape B respectively:

Sequence #1 = BLMALSCMM
Sequence #2 = AMLALM

So M = 9 and N = 6 (the length of sequence #1 and sequence #2, respectively)

There are three steps for compute a similarity between 2 strings.
1. Initialization
2. Matrix (scoring)
3. Trace back and alignment.

*A. Initialization Step*

The first step in the global alignment dynamic programming approach is to create a matrix with M+ 1 columns and N+ 1 rows where M and N correspond to the size of the sequences to be aligned. The values of the matrix are initialized by 0.(Fig 7)

*B. Matrix (scoring)*

The computing starts in the upper left hand corner in the matrix and finds the maximal score $F_{i,j}$ for each position in the matrix. $F_{i,j}$ is calculated using the formula (5)

$F_{i,j}$ = MAX[
  $F_{i-1, j-1}$ + $S_{i,j}$ (match/mismatch in the diagonal),
  $F_{i,j-1}$ + w (gap in sequence#1),      (5)
  $F_{i-1,j}$ + w (gap in sequence#2)
]

**F** represents the score for the matrix position. **W** represents a gap of penalty score, and its value equal to "-2". **S** represents the match/mismatch score at the diagonal position, and its value is defined as following:

- Value 2 for matching.
- A large weight (with value lower than one) for the substitution of two adjacent symbols: for example, the score between A and B was take 1, but the score between A and C equal to 1/2 (0.5) because there are not adjacent, and similarly the score between S and M or M and L was taken to be equal to 1.
- (-2) for others. (more details see Fig. 8)

|   | B | L | M | A | L | S | C | M | M |
|---|---|---|---|---|---|---|---|---|---|
|   | 0 | 0 | 0 | 0 | 0 | 0 | 0 | 0 | 0 |
| A | 0 | 0 | 0 | 0 | 0 | 0 | 0 | 0 | 0 |
| M | 0 | 0 | 0 | 0 | 0 | 0 | 0 | 0 | 0 |
| L | 0 | 0 | 0 | 0 | 0 | 0 | 0 | 0 | 0 |
| A | 0 | 0 | 0 | 0 | 0 | 0 | 0 | 0 | 0 |
| L | 0 | 0 | 0 | 0 | 0 | 0 | 0 | 0 | 0 |
| M | 0 | 0 | 0 | 0 | 0 | 0 | 0 | 0 | 0 |

Fig.7: Initialization Step

|   | A | B | C | D | E | F | S | M | L |
|---|---|---|---|---|---|---|---|---|---|
| A | 2 | 1 | 1/2 | 1/3 | 1/4 | 1/5 | -2 | -2 | -2 |
| B | 1 | 2 | 1 | 1/2 | 1/3 | 1/4 | -2 | -2 | -2 |
| C | 1/2 | 1 | 2 | 1 | 1/2 | 1/3 | -2 | -2 | -2 |
| D | 1/3 | 1/2 | 1 | 2 | 1 | 1/2 | -2 | -2 | -2 |
| E | 1/4 | 1/3 | 1/2 | 1 | 2 | 1 | -2 | -2 | -2 |
| F | 1/5 | 1/4 | 1/3 | 1/2 | 1 | 2 | -2 | -2 | -2 |
| S | -2 | -2 | -2 | -2 | 1/2 | -2 | 2 | 1 | 1/2 |
| M | -2 | -2 | -2 | -2 | -2 | -2 | 1 | 2 | 1 |
| L | -2 | -2 | -2 | -2 | -2 | -2 | 1/2 | 1 | 2 |

**Fig 8**. the score between different symbols.

|   |   | B | L | M | A | L | S | C | M | M |
|---|---|---|---|---|---|---|---|---|---|---|
|   | 0 | 0 | 0 | 0 | 0 | 0 | 0 | 0 | 0 | 0 |
| A | 0 | 1 | -1 | -2 | 2 | 0 | -2 | 0,5 | -1,5 | 0 |
| M | 0 | -1 | 2 | 1 | 0 | 3 | 1 | -1 | 2,5 | 0,5 |
| L | 0 | -2 | 1 | 3 | 1 | 2 | 3,5 | 1,5 | 0,5 | 3,5 |
| A | 0 | 1 | -1 | 1 | 5 | 3 | 1,5 | 4 | 2 | 1,5 |
| L | 0 | -1 | 3 | 1 | 3 | 7 | 5 | 3 | 5 | 3 |
| M | 0 | -2 | 1 | 5 | 3 | 5 | 8 | 5 | 5 | 7 |

**Fig 9**. Matrix scoring.

|   |   | B | L | M | A | L | S | C | M | M |
|---|---|---|---|---|---|---|---|---|---|---|
|   | 0 | 0 | 0 | 0 | 0 | 0 | 0 | 0 | 0 | 0 |
| A | 0 | 1 | -1 | -2 | 2 | 0 | -2 | 0,5 | -1,5 | 0 |
| M | 0 | -1 | 2 | 1 | 0 | 3 | 1 | -1 | 2,5 |   |
| L | 0 | -2 | 1 | 3 | 1 | 2 | 3,5 | 1,5 | 0,5 | 3,5 |
| A | 0 | 1 | -1 | 1 | 5 | 3 | 1,5 | 4 | 2 | 1,5 |
| L | 0 | -1 | 3 | 1 | 3 | 7 | 5 | 3 | 5 | 3 |
| M | 0 | -2 | 1 | 5 | 3 | 5 | 8 | 5 | 5 | 7 |

**Fig.10**: Matrix traceback

## C. Traceback and optimal cost.

After filling in all of the values, the score matrix is represented in Fig.9 and the maximum alignment score for the two test sequences is 7. The traceback step determines the actual alignment(s) that result in the maximum score. Note that with a simple scoring algorithm such as one that is used here, there are likely to be multiple maximal alignments.

Fig. 10 shows the path of the matrix traceback. The Traceback step begins in the M,N position in the matrix. Traceback takes the current cell and looks to the neighbor cells that could be direct predecessors. This means it looks to the neighbor to the left (gap in sequence #2), the diagonal neighbor (match/mismatch), and the neighbor above it (gap in sequence #1). The algorithm for traceback chooses as the next cell in the sequence one of the possible predecessor.

After this step, two string sequences are aligned as follows:

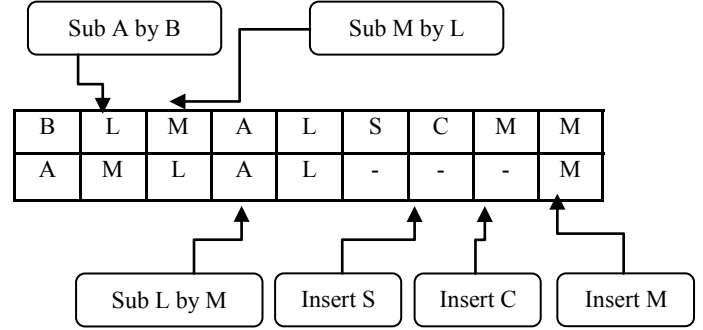

The approach of dynamic programming allows you to find the optimal alignment between sequences of strings of different lengths. Since the algorithm can handle the alignment of different string lengths. Therefore, the algorithm can calculate a large number of comparisons. The string sequences are aligned for higher scores. The system can find similar sequences against a query sequence by the sort key alignment score in ascending order. Since each strong of sequence is a set of points, the system may display similar sets of parts of the two shapes as search results.

## IV. EXPERIMENTS AND RESULTS

The proposed method is tested on the database of MPEG7 CE Shape-1 Part B. The database consists of 70 classes and 20 images per category in total 1,400 images. Figure 11 shows some examples. In a test the system extracted 20 shapes whose scores are higher are extracted from the database.

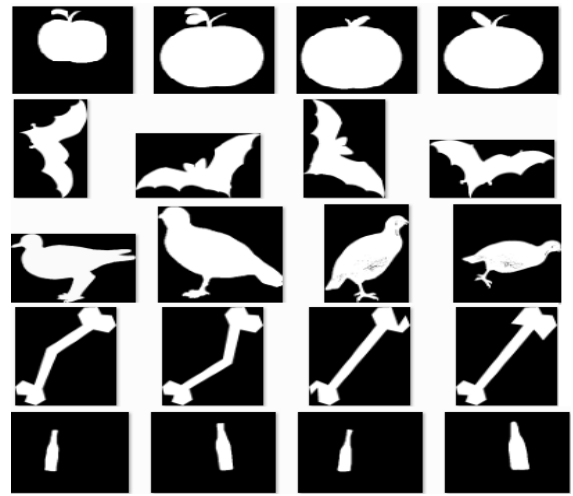

**Fig.11** Exemplar shapes in the MPEG-7 shape database for five different categories.

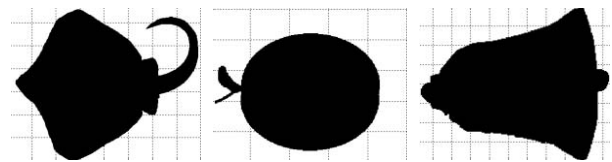

**Fig. 12**: Input shapes.

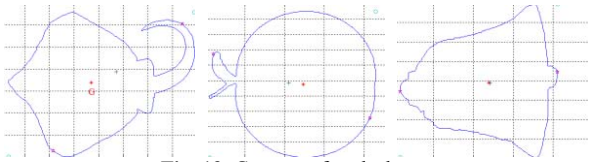
**Fig. 13**: Contour of each shape.

| Method/Algorithm | Retrieval Score(%) | Recognition Score(%) |
|---|---|---|
| Shape Context [9] | 76.51 | --- |
| Curve Edit Distance [16] | 78.1 | --- |
| Curvature Scale Space [17] | 81.12 | --- |
| Polygonal multi-Resolution [18] | 84.33 | 97,57 |
| String of Symbols [19] | 82 | 96.8 |
| Symbolic Representation [6] | 83.45 | 97.57 |
| HPM-Fn | 86.35 | --- |
| Ours | 89.743 | 96.862 |

**Fig. 14**: Comparison of results for different algorithms tested on the MPEG-7 database.

Input shapes are presented as shown in Fig. 12; the system detects the first time the contour of each shape (Fig. 13) and then applies the algorithm to retrieve similar shapes

Results are expressed as a percentage. The table in Fig. 14 shows the score of retrieval and recognition using the proposed technique. The results are compared with some old methods and techniques in MPEG-7.

For recovery, our algorithm was compared with known solutions and is advanced with a small difference compared to other solutions. However, for recognition our algorithm outperforms some previous approaches.

THE OCCLUSIONS :

In our treatment of occlusions we took some completes shapes which were used as reference images for experimentation. The percentage of matches between the reference shapes and other shapes is obtained, and we computed the percentage of recognition for different values of k (k is the number of angles quantified for each state, return to Section I-C), Fig 15 shows the result of recognition between the shapes according to the percentage of occlusion and value of k.

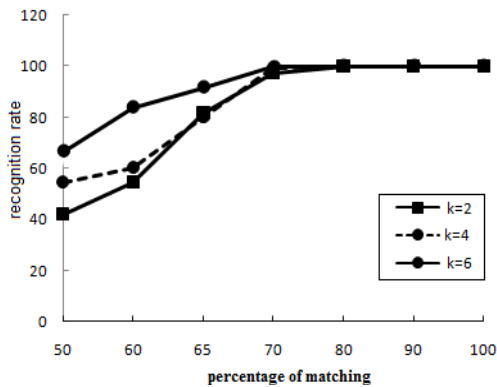

**Fig 15**: Recognition result between the shapes according to k.

## V. CONCLUSION

We have presented a new approach to finding the best matching between two shapes using the technique of dynamic programming. We developed a fast and efficient algorithm to find a good similarity between the shapes. The contours and alignment using the technique of shape context and dynamic programming are the keys to our method. We demonstrated the superiority of our approach over traditional approaches in databases as MPEG-7.